\documentclass{llncs}
\usepackage{makeidx}  
\usepackage{times} 
\usepackage{helvet}
\usepackage{courier}
\usepackage[hyphens]{url} 
\usepackage{color}

\usepackage{graphicx}
\graphicspath{{figs/}}
\usepackage[square,sort,comma,numbers]{natbib}
\usepackage[caption=false]{subfig}
\usepackage{amsmath, mathtools, amsfonts}
\usepackage{booktabs}
\usepackage{xcolor}
\usepackage{cleveref}

\begin{document}
\frontmatter          
\pagestyle{headings}  
\addtocmark{Learning Sequential Latent Variable Models from Multimodal Time Series Data} 
\mainmatter              
\title{Learning Sequential Latent Variable Models\\ from Multimodal Time Series Data}
\titlerunning{Learning Sequential Latent Variable Models from Multimodal Time Series Data}  
%
\author{Oliver Limoyo\inst{1,*} \and Trevor Ablett\inst{1} \and Jonathan Kelly\inst{1}}
\authorrunning{Oliver Limoyo et al.} 
%
\tocauthor{Oliver Limoyo, Trevor Ablett, and Jonathan Kelly}
\institute{$^{1}$ University of Toronto Institute for Aerospace Studies (UTIAS), Toronto, Canada\\
    \tt\small <first name>.<last name>@robotics.utias.utoronto.ca\\
    $^{*}$Corresponding author}

\maketitle              

\begin{abstract}
  Sequential modelling of high-dimensional data is an important problem that appears in many domains including model-based reinforcement learning and dynamics identification for control.
  Latent variable models applied to sequential data (i.e., latent dynamics models) have been shown to be a particularly effective probabilistic approach to solve this problem, especially when dealing with images.
  However, in many application areas (e.g., robotics), information from multiple sensing modalities is available---existing latent dynamics methods have not yet been extended to effectively make use of such multimodal sequential data.
  Multimodal sensor streams can be correlated in a useful manner and often contain complementary information across modalities.
  In this work, we present a self-supervised generative modelling framework to jointly learn a probabilistic latent state representation of multimodal data and the respective dynamics. 
  Using synthetic and real-world datasets from a multimodal robotic planar pushing task, we demonstrate that our approach leads to significant improvements in prediction and representation quality.
  Furthermore, we compare to the common learning baseline of concatenating each modality in the latent space and show that our principled probabilistic formulation performs better.
  Finally, despite being fully self-supervised, we demonstrate that our method is nearly as effective as an existing supervised approach that relies on ground truth labels.
  
\end{abstract}

\section{Introduction}
\label{intr}
\noindent Sequential modelling of high-dimensional data is a challenging problem that is encountered in many domains such as visual model-based reinforcement learning \cite{Hafner2019-jj} and image-based dynamics identification for control \cite{Watter2015-ge}.
Recently, there has been broad interest in studying the problem through the lens of generative latent variable models. 
These methods embed high-dimensional observations into a lower-dimensional representation space, or \textit{latent space}, often by means of a variational autoencoder (VAE) \cite{Kingma2013-kw, Rezende2014-ug}, where the (latent) dynamics can be identified in a self-supervised manner.
Importantly, latent dynamics models form the backbone of many recent methods for image-based control, reinforcement learning \cite{hafner2021mastering, rafailov2021offline}, and motion planning with complex robotic systems \cite{ichter2019robot, lippi2020latent}.
When dealing with images as a single modality, this approach has been shown to be particularly effective in various existing works \cite{zhang2019solar, Wahlstrom2015-wd, Watter2015-ge, Karl2016-ok, FraccaroKPW17}.
However, in many robotic systems, images are accompanied by data from multiple additional sensing modalities with varying characteristics, that is, as \textit{multimodal} data.
These multimodal data may contain complementary information, patterns, and useful statistical correlations across modalities. 
Prior research has demonstrated the advantages of capturing multimodal information in the contexts of classification and regression of multimedia data \cite{ngiam2011, 8269806}, learning from demonstration \cite{hristov2021learning}, localization \cite{14425}, and control policy learning \cite{quteprints109664}.
Conversely, the use of multimodal data has not yet been studied in the context of learned latent dynamics models.  

In this paper, we present a novel probabilistic framework for learning latent dynamics from multimodal time series data.
Inspired by the multimodal variational autoencoder (MVAE) architecture \cite{wu2018}, we employ a product of experts \cite{poe2002} to encode all data modalities into a shared probabilistic latent representation while jointly learning the dynamics in a self-supervised manner with a recurrent neural network (RNN) \cite {cho2014properties}.

\begin{figure}[t]
	\centering
	\includegraphics[width=0.99\textwidth]{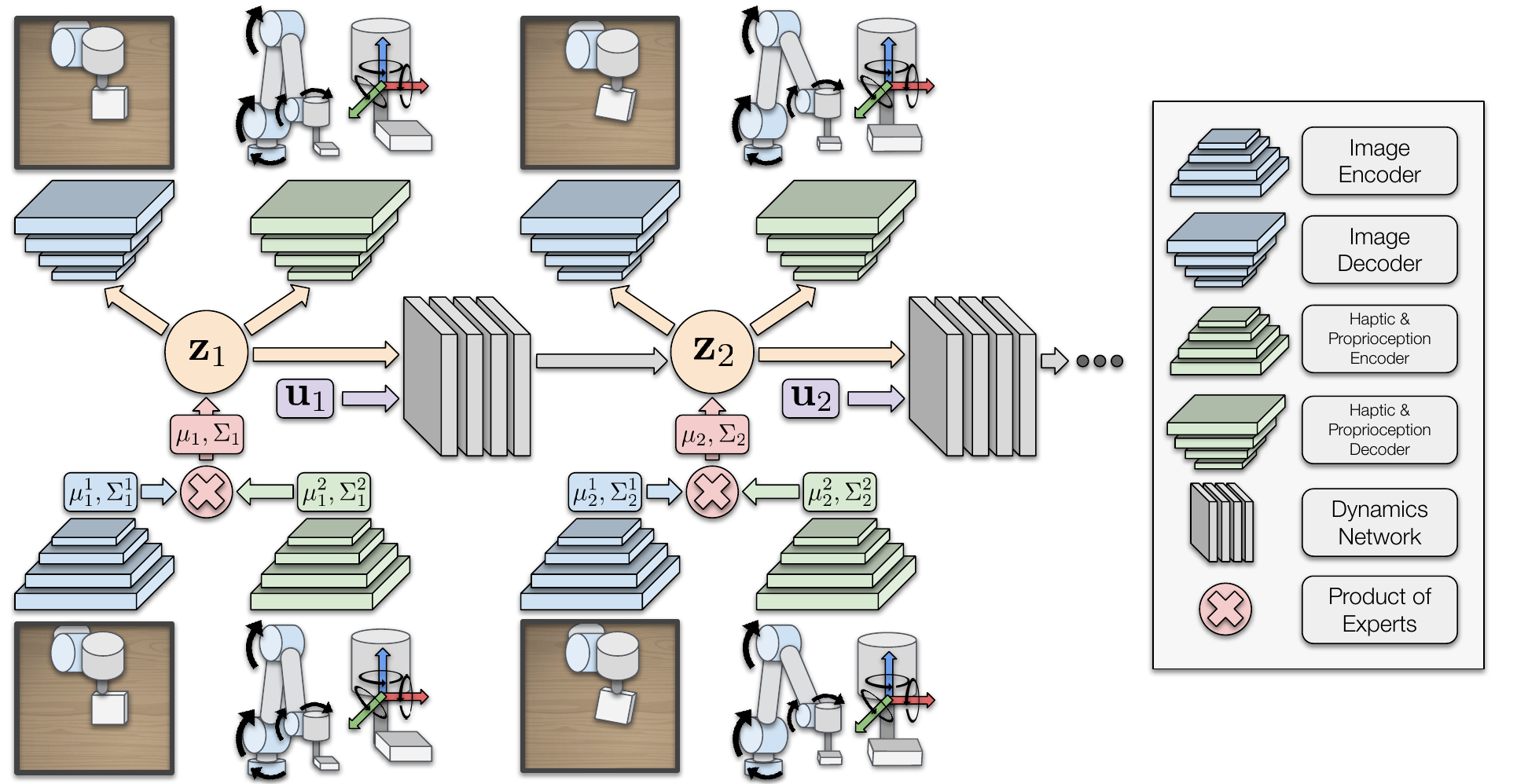}
	\caption{We learn a sequential latent variable model for multimodal time series data (e.g., image, proprioceptive, and haptic data). Each modality is separately encoded into a probabilistic latent distribution, in this case a Gaussian distribution parametrized by $\boldsymbol{\mu}$ and $\mathbf{\Sigma}$, and then combined into a joint representation using a product of experts. Simultaneously, we learn the respective (latent) dynamics based on the control inputs \textbf{u}.}
	\label{fig:overview} 
	\vspace{-5mm}
\end{figure}

For validation and demonstration purposes, we use simulated and real-world datasets collected from a robotic manipulator pushing task involving three heterogenous data modalities: images from a camera, force and torque readings from a force-torque sensor, and proprioceptive readings from the manipulator encoders. 
Figure \ref{fig:overview} provides a visual summary of our approach. 
Our main contributions are as follows:  
\begin{enumerate}
	\item a formulation of a variational inference model for self-supervised training of latent dynamics models specifically with multimodal time series data;
	\item experimental results demonstrating that our method of incorporating multimodal data in the latent dynamics framework has superior representative and predictive capability when compared to the baseline of simply concatenating each modality \cite{kiela2014learning, d2015review, regneri2013grounding, gandhi2020swoosh};
	\item experimental results demonstrating that our self-supervised method achieves results comparable to an existing supervised method \cite{lee2020}, which requires ground-truth labels, when used to capture task-relevant state information and dynamics; and
	\item an open-source implementation of our method and experiments in PyTorch \cite{NEURIPS2019_9015}.\footnote{\url{https://github.com/utiasSTARS/visual-haptic-dynamics}} 
\end{enumerate}

\section{Related Work}
\label{rw}
In this section, we survey papers related to the modelling of high-dimensional sequential data with latent variable models and learning with multimodal data in machine learning and robotics. 
We pay particular attention to vision as it is one of the most commonly-encountered modalities across many domains. 

\vspace{-\baselineskip}
\subsubsection{Latent Dynamics}
Early deep dynamical models \cite{Wahlstrom2015-wd} used the bottleneck of a standard autoencoder as the compressed state from which the dynamics were learned in a tractable manner.
Probabilistic extensions using the VAE were studied later
\cite{Watter2015-ge, Banijamali2017-nx}.
A significant amount of work has been carried out with sequential image data in the context of learned probabilistic state space models \cite{Krishnan2016-wz, Karl2016-ok} and differentiable filtering \cite{NIPS2016_697e382c}. 
These approaches attempt to combine the structure and interpretability of probabilistic graphical models with the flexibility and representational capacity of neural networks.
Probabilistic graphical models have been used as a way to impose structure for fast and exact inference \cite{Johnson2016-yr, FraccaroKPW17} and to filter out novel or out-of-distribution images using a notion of uncertainty that comes from generative models \cite{2020_Limoyo_Heteroscedastic}.
The closely-related topic of image-based transition models, or \textit{world models}, has been studied in the reinforcement learning literature \cite{Hafner2019-jj, Ha2018-ho}. 
Our work is an extension to these latent dynamics models that makes them amenable to multimodal data.

\vspace{-\baselineskip}
\subsubsection{Multimodal Machine Learning}
Machine learning has been applied to the problem of learning representations and patterns of multimodal data for various downstream tasks. 
A good summary of the existing literature focused on applications involving multimedia data (e.g., video, text, and audio) is provided in \cite{8269806}. 
Probabilistic methods have also been applied to model the joint and conditional distributions of non-sequential multimodal data.
Examples of these include the Restricted Boltzman Machine (RBM) \cite{ngiam2011, srivastava2014} and the VAE \cite{wu2018, suzuki2016joint, VedantamFH018}. 
Our work is most similar to the latter of these two approaches. We build upon the MVAE \cite{wu2018}, but, critically, we apply and extend the framework to the sequential setting so that it is amenable to the application of capturing the latent dynamics of multimodal data.

\vspace{-\baselineskip}
\subsubsection{Multimodal Learning for Robotics}
Our work is most similar in nature to \cite{gandhi2020swoosh}, where the authors use audio data to augment a deterministic state-based forward (or dynamics) model. 
However, in \cite{gandhi2020swoosh}, the audio data are taken from a previous random interaction and do not provide causally-related information to the forward model---the audio data is simply used to augment the representation.
In contrast, we directly model the dynamics of the observed multimodal data.
We also do not assume a relaxed deterministic state-based setting and instead learn a probabilistic representation from raw multimodal data directly (as opposed to dealing with difficult-to-acquire state labels).

Similar to the approach presented in this paper, other groups have investigated the use of learned differentiable filters with multimodal measurement models of vision, proprioceptive, and haptic data, relying on ground-truth annotations \cite{lee2020}.
However, in many cases ground-truth labels are expensive or impossible to acquire, which may hinder the scalability of such methods. 
We leverage recent work in variational inference and devise a self-supervised generative approach to bypass this limitation.
We do so by maximizing a proper lower bound of the marginal likelihood of the data itself.

Other works have also leveraged the MVAE architecture for learned localization with multimodal data \cite{14425}.
Closer to our work, the authors of \cite{rezaei2021learning} demonstrate a technique to learn a notion of `intuitive physics' in a self-supervised manner by applying the MVAE architecture as a generative model of multimodal sensor measurements.
Specifically, future sensor measurements resulting from interaction with objects are decoded based on an encoding of the current sensor measurements.
In our work, as opposed to directly decoding future transitions, we learn a dynamics model based on the compressed latent space; we therefore have the choice to predict while remaining in a low-dimensional space and without having to decode, which saves a significant amount of computation and memory.

\section{Methodology}
\label{me}
We begin by presenting a baseline sequential latent variable model in Section \ref{slv} 
and then demonstrate how to extend this framework to multimodal data in Section \ref{mslv}. 

\subsection{Sequential Latent Variable Models}
\label{slv}
We consider a sequence of observations $\mathbf{x} = \{\mathbf{x}_{t}\}_{t=1}^{T}$ of a single modality with respective control inputs $\mathbf{u} = \{\mathbf{u}_{t}\}_{t=1}^{T-1}$.
We then introduce latent variables $\mathbf{z} = \{\mathbf{z}_{t}\}_{t=1}^{T}$ to create a joint distribution $p_{\mathbf{\theta}}(\mathbf{x}, \mathbf{z} | \mathbf{u})$, where $\mathbf{\theta}$ are the learnable parameters of our distributions that are parametrized by neural networks.
We factorize the joint distribution of the generative process as $p_{\mathbf{\theta}}(\mathbf{x}, \mathbf{z}| \mathbf{u}) = p_{\mathbf{\theta}}(\mathbf{x} | \mathbf{z}, \mathbf{u})\,p_{\mathbf{\theta}}(\mathbf{z} | \mathbf{u})$, where:
\begin{equation} \label{eq:transition}
p_{\mathbf{\theta}}(\mathbf{z} | \mathbf{u}) = p(\mathbf{z}_{1})\prod_{t=2}^{T}p_{\mathbf{\theta}}(\mathbf{z}_{t} | \mathbf{z}_{t-1}, \mathbf{u}_{t-1}),
\end{equation}
and 
\begin{equation} \label{eq:emission}
p_{\mathbf{\theta}}(\mathbf{x} | \mathbf{z}, \mathbf{u}) = \prod_{t=1}^{T} p_{\mathbf{\theta}}(\mathbf{x}_{t} | \mathbf{z}_{t}).
\end{equation}
We model the latent dynamics with the distribution $p_{\mathbf{\theta}}(\mathbf{z}_{t} | \mathbf{z}_{t-1}, \mathbf{u}_{t-1})$ and the observation or measurement model with $p_{\mathbf{\theta}}(\mathbf{x}_{t} | \mathbf{z}_{t})$. The distribution $p(\mathbf{z}_{1})$ is an arbitrary initial distribution with high uncertainty. 
The goal of learning is to maximize the marginal likelihood of the data or the evidence, given for a single sequence as
\begin{equation} \label{eq:marginal}
p_{\mathbf{\theta}}(\mathbf{x} | \mathbf{u}) = \int p_{\mathbf{\theta}}(\mathbf{x} | \mathbf{z}, \mathbf{u})\,p_{\mathbf{\theta}}(\mathbf{z} | \mathbf{u})\text{d}\mathbf{z},
\end{equation}
with respect to the parameters $\mathbf{\theta}$.
Unfortunately, in the general case with this model, the posterior distribution used for inference, $p(\mathbf{z}| \mathbf{x}, \mathbf{u})$, is intractable. 
A common solution from recent work in variational inference \cite{Kingma2013-kw, Rezende2014-ug} is to introduce a recognition or inference model $q_{\mathbf{\phi}}(\mathbf{z}| \mathbf{x}, \mathbf{u})$ with parameters $\mathbf{\phi}$ to approximate the intractable posterior. 
This leads to the following lower bound on the marginal log-likelihood or the evidence lower bound (ELBO):
\begin{equation}
\label{eq:elbo}
\ln p_{\theta}(\mathbf{x} | \mathbf{u}) \geq \mathbb{E}_{q_{\phi}(\mathbf{z}|\mathbf{x}, \mathbf{u})}[\ln p_{\theta}(\mathbf{x}|\mathbf{z}, \mathbf{u})] - \text{KL}(q_{\phi}(\mathbf{z}|\mathbf{x},\mathbf{u}) \| p_{\theta}(\mathbf{z}|\mathbf{u})).
\end{equation}
Maximizing this lower bound can be shown to be equivalent to minimizing the KL-divergence between the true posterior $p_{\mathbf{\theta}}(\mathbf{z}| \mathbf{x}, \mathbf{u})$ and the recognition model $q_{\mathbf{\phi}}(\mathbf{z}| \mathbf{x}, \mathbf{u})$. 
The resulting optimization objective, as denoted by Eq. \ref{eq:elbo}, is based on an expectation with respect to the distribution $q_{\mathbf{\phi}}(\mathbf{z}| \mathbf{x}, \mathbf{u})$, which itself is based on the parameters $\mathbf{\phi}$. 
As is typically done, we restrict $q_{\mathbf{\phi}}(\mathbf{z}| \mathbf{x}, \mathbf{u})$ to be a Gaussian variational approximation. 
This enables us to use stochastic gradient descent (i.e., using Monte Carlo estimates of the gradient) via the reparameterization trick \cite{Kingma2013-kw} to optimize the lower bound. 
The specific choice for the factorization of $q_{\mathbf{\phi}}(\mathbf{z}| \mathbf{x}, \mathbf{u})$ varies depending on the application (i.e., prediction, smoothing, or filtering).
Given our intended applications of prediction we choose to only use causal (i.e., current and past) information for inference,
\begin{equation} \label{eq:posterior}
q_{\mathbf{\phi}}(\mathbf{z}| \mathbf{x}, \mathbf{u}) = \prod_{t=1}^{T} q_{\mathbf{\phi}}(\mathbf{z}_{t}| \mathbf{x}_{\leq t}, \mathbf{u}_{< t}).
\end{equation}

\subsection{Multimodal Sequential Latent Variable Models}
\label{mslv}
We now extend and generalize the sequential latent variable model defined above to the multimodal case. 
We consider N sequences of separate observations or modalities, $\mathbf{X} = \{\mathbf{x}^{n}\}_{n=1}^{N}$, where we assume that each sequence is of equivalent length $T$: $\mathbf{x}^{n} = \{\mathbf{x}^{n}_{t}\}_{t=1}^{T}$. 
As done in the previous section, we include the respective control inputs $\mathbf{u} = \{\mathbf{u}_{t}\}_{t=1}^{T-1}$ and again introduce a set of latent variables $\mathbf{z} = \{\mathbf{z}_{t}\}_{t=1}^{T}$ as a joint lower-dimensional latent space containing some underlying dynamics of interest.
The final joint distribution factorizes as $p_{\mathbf{\theta}}(\mathbf{X}, \mathbf{z}| \mathbf{u}) = p_{\mathbf{\theta}}(\mathbf{X}| \mathbf{z}, \mathbf{u})\, p_{\mathbf{\theta}}(\mathbf{z} | \mathbf{u})$. 
We choose to define the generative process as follows:
\begin{equation} \label{eq:emission_mm}
\begin{split}
p_{\mathbf{\theta}}(\mathbf{X}| \mathbf{z}, \mathbf{u}) &= \prod_{n=1}^{N} p_{\mathbf{\theta}}(\mathbf{x}^{n} | \mathbf{z}, \mathbf{u}) \\
&= \prod_{n=1}^{N} \prod_{t=1}^{T} p_{\mathbf{\theta}}(\mathbf{x}_{t}^{n} | \mathbf{z}_{t}),
\end{split}
\end{equation}
and with $p_{\mathbf{\theta}}(\mathbf{z} | \mathbf{u})$ remaining the same as shown in Eq. \ref{eq:transition}. 
The marginal likelihood is then
\begin{equation} \label{eq:marginal_mm}
p_{\mathbf{\theta}}(\mathbf{X} | \mathbf{u}) = \int p_{\mathbf{\theta}}(\mathbf{X} | \mathbf{z}, \mathbf{u})\, p_{\mathbf{\theta}}(\mathbf{z} | \mathbf{u})\text{d}\mathbf{z},
\end{equation}
and the respective ELBO, for a single sequence, is then: 
\begin{multline} \label{eq:elbo_mm}
\ln p_{\theta}(\mathbf{X} | \mathbf{u}) \geq \mathbb{E}_{q_{\phi}(\mathbf{z}|\mathbf{X}, \mathbf{u})}[\sum_{\mathbf{x}^{n} \in \mathbf{X}}\ln p_{\theta}(\mathbf{x}^{n}|\mathbf{z}, \mathbf{u})] - \text{KL}(q_{\phi}(\mathbf{z}|\mathbf{X},\mathbf{u}) \| p_{\mathbf{\theta}}(\mathbf{z}|\mathbf{u})).
\end{multline}
In order to decide on the factorization of $q_{\phi}(\mathbf{z}|\mathbf{X},\mathbf{u})$, we draw inspiration from the MVAE architecture \cite{wu2018}, and base our inference network on the structure of the true multimodal posterior $p(\mathbf{z}| \mathbf{X}, \mathbf{u})$:
\begin{equation} \label{eq:inference_mm_factorization}
\begin{split}
p(\mathbf{z}| \mathbf{X}, \mathbf{u}) &= \frac{p(\mathbf{z} | \mathbf{u})\, p(\mathbf{X} | \mathbf{z}, \mathbf{u})}{p(\mathbf{X}| \mathbf{u})} \\
&= \frac{p(\mathbf{z} | \mathbf{u})}{p(\mathbf{x}^{1}, ..., \mathbf{x}^{N}| \mathbf{u})} \prod_{n=1}^{N} p(\mathbf{x}^{n} | \mathbf{z}, \mathbf{u}) \\
&= \frac{p(\mathbf{z} | \mathbf{u})}{p(\mathbf{x}^{1}, ..., \mathbf{x}^{N}| \mathbf{u})} \prod_{n=1}^{N} \frac{p(\mathbf{z} | \mathbf{x}^{n}, \mathbf{u}) p(\mathbf{x}^{n} | \mathbf{u})}{p(\mathbf{z}|\mathbf{u})} \\
&= \frac{\prod_{n=1}^{N}p(\mathbf{x}^{n} | \mathbf{u})}{p(\mathbf{x}^{1}, ..., \mathbf{x}^{N}| \mathbf{u})} \cdot \frac{\prod_{n=1}^{N} p(\mathbf{z} | \mathbf{x}^{n}, \mathbf{u})}{\prod_{n=1}^{N-1} p(\mathbf{z}|\mathbf{u})} \\
&\propto \frac{\prod_{n=1}^{N} p(\mathbf{z} | \mathbf{x}^{n}, \mathbf{u})}{\prod_{n=1}^{N-1} p(\mathbf{z}|\mathbf{u})}.
\end{split}
\end{equation}
Based on the last term in Eq. \ref{eq:inference_mm_factorization}, the final factorization of the joint posterior is then a quotient between a product of the individual modality-specific posteriors and the prior. 
Accordingly, we choose our inference model to be $q(\mathbf{z} | \mathbf{X}, \mathbf{u}) = \frac{\prod_{n=1}^{N} q(\mathbf{z} | \mathbf{x}^{n}, \mathbf{u})}{\prod_{n=1}^{N-1} p(\mathbf{z}|\mathbf{u})}$. 
We also use the same representation reformulation trick from the MVAE \cite{wu2018} and set $q(\mathbf{z} | \mathbf{x}^{n}, \mathbf{u}) = \tilde{q}(\mathbf{z} | \mathbf{x}^{n}, \mathbf{u})\, p(\mathbf{z}|\mathbf{u})$ in order to produce a simpler and numerically more stable product of experts,
\begin{equation} \label{eq:inference_mm_poe}
\begin{split}
q_{\mathbf{\phi}}(\mathbf{z} | \mathbf{X}, \mathbf{u}) = p_{\mathbf{\theta}}(\mathbf{z}|\mathbf{u})\prod_{n=1}^{N} \tilde{q}_{\mathbf{\phi}}(\mathbf{z} | \mathbf{x}^{n}, \mathbf{u}),
\end{split}
\end{equation}
where $\tilde{q}_{\mathbf{\phi}}(\mathbf{z} | \mathbf{x}^{n}, \mathbf{u})$ would be equivalent to the standard single modality posterior shown in Eq. \ref{eq:posterior}. 
Finally, we can further factorize the posterior in Eq. \ref{eq:inference_mm_poe} into a more intuitive form for our sequential setting:
\begin{align} \label{eq:inference_mm_poe_fact}
\begin{split}
    q_{\mathbf{\phi}}(\mathbf{z} | \mathbf{X}, \mathbf{u}){}& = \underbrace{p(\mathbf{z}_{1})\prod_{n=1}^{N} \tilde{q}_{\mathbf{\phi}}(\mathbf{z}_{1}| \mathbf{x}^{n}_{1})}_{q_{\mathbf{\phi}}(\mathbf{z}_{1} | \mathbf{x}_{1}^{1}, ..., \mathbf{x}_{1}^{N}) }
        \prod_{t=2}^{T} \underbrace{\left( p_{\mathbf{\theta}}(\mathbf{z}_{t} | \mathbf{z}_{t-1}, \mathbf{u}_{t-1})\prod_{n=1}^{N} \tilde{q}_{\mathbf{\phi}}(\mathbf{z}_{t}| \mathbf{x}^{n}_{\leq t}, \mathbf{u}_{< t}) \right)}_{q_{\mathbf{\phi}}(\mathbf{z}_{t} | \mathbf{x}_{\leq t}^{1}, ..., \mathbf{x}_{\leq t}^{N}, \mathbf{u}_{< t})} \nonumber
\end{split}\\
    {}& = \prod_{t=1}^{T} q_{\mathbf{\phi}}(\mathbf{z}_{t} | \mathbf{x}_{\leq t}^{1}, ..., \mathbf{x}_{\leq t}^{N}, \mathbf{u}_{< t}).
\end{align}
Our factorization reveals that, at every time step of Eq. \ref{eq:inference_mm_poe_fact}, each data modality is first separately encoded into a Gaussian distribution by its own inference model.
A product is then taken of each modality-specific distributions and the prior, which is also the transition distribution (i.e., the dynamics) of our latent space. Interestingly, we recover a similar form to the commonly-used recurrent state space model \cite{Hafner2019-jj}, where the transition distribution is included in the posterior.
The product of distributions at each time step are not generally solvable in closed form. 
However, by assuming Gaussian distributions for the prior dynamics and each modality-specific inference distribution, we end up with a final product of Gaussians for which a closed-form analytical solution does exist. 
Conveniently, a product of Gaussians is also itself a Gaussian \cite{cao2014generalized}.

\section{Experiments}
\label{exp}
In this section, we present implementation details and an empirical evaluation of our method.
For our experiments, we chose to study a common multimodal task: planar pushing with a robotic manipulator using vision, haptic, and proprioceptive data from a camera, a force-torque sensor, and joint encoders, respectively. 
\begin{figure}
	\vspace*{-4mm}
    \centering
    \subfloat[PyBullet pushing simulation.\label{fig:datasetup_sim}]{
        \includegraphics[height=1.75in]{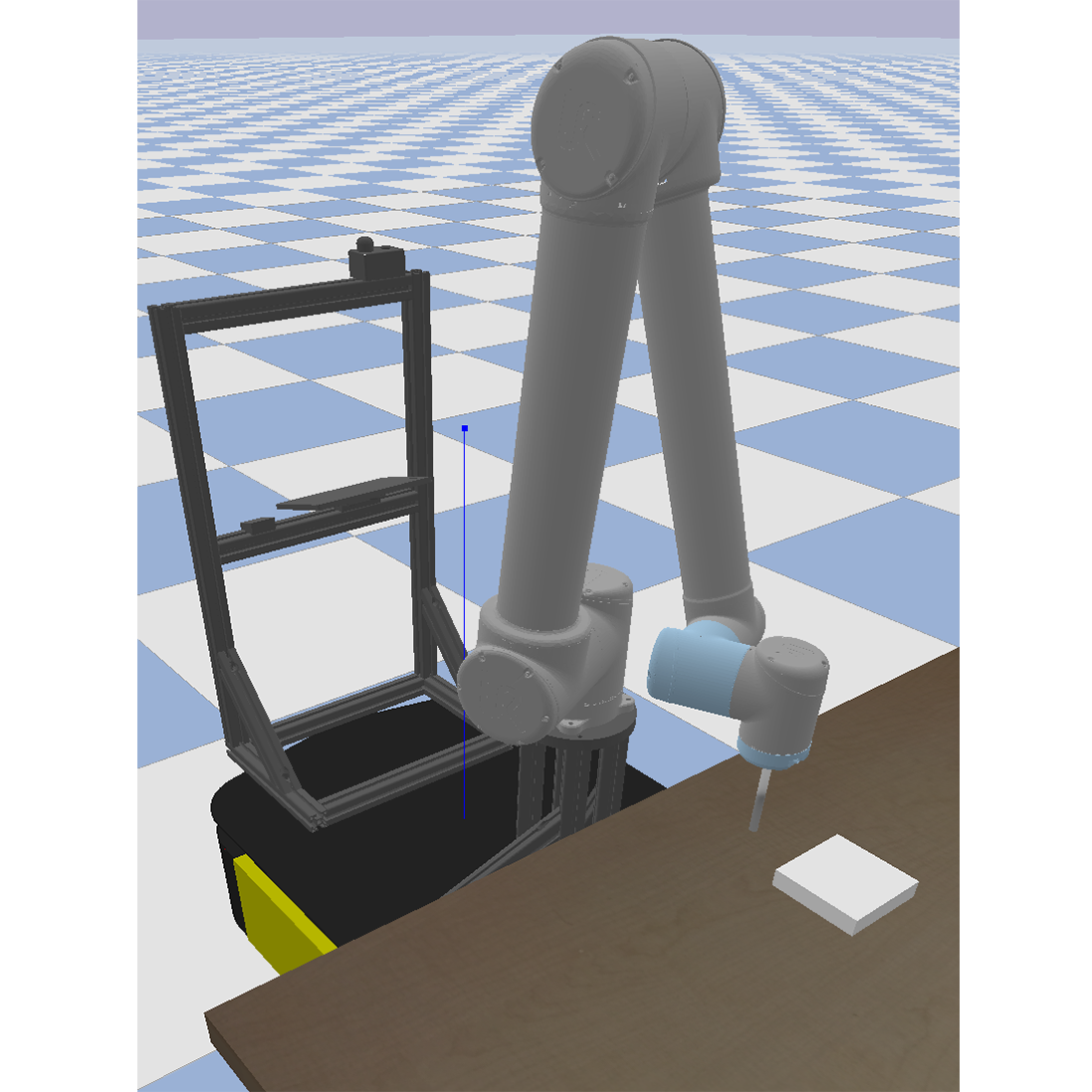}    
    }\hfil
    \subfloat[Real-world pushing task.\label{fig:datasetup_real}]{
        \includegraphics[height=1.75in]{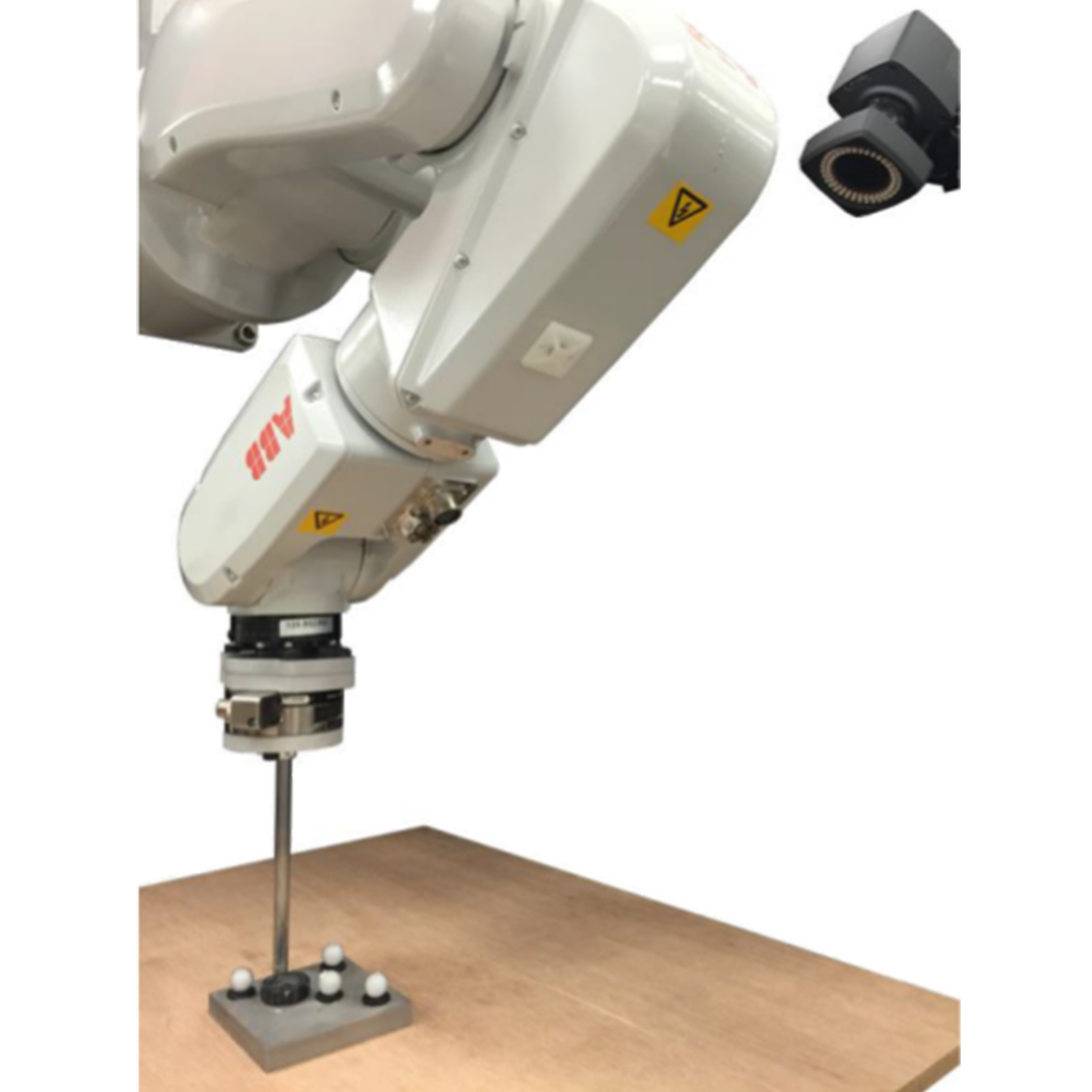}    
    }
    \caption{Environments from which the datasets were collected. Real-world setup image from existing MIT pushing dataset \cite{yu2016push}.}
	\label{fig:datasetup}
	\vspace{-5mm}
\end{figure}
Planar pushing involves complex contact dynamics that are difficult to model with vision alone, while the multimodal sensor data produced are highly heterogeneous in dimension and quality.

We compared five different models to study the effects of multimodality on this type of task.
The models were, in order: 1) vision-only (denoted by V), 2) visual-proprioceptive (denoted by VP), 3) visual-haptic (denoted by VH), 4) visual-haptic-proprioceptive (denoted by VHP), and 5) a commonly-used baseline where the latent embedding of each data modality is simply concatenated (denoted by VHP-C).

\subsection{Datasets}
\label{da}
We use both a synthetic and a real-world dataset as shown in Figure \ref{fig:datasetup}. We provide more details on the data and the collection procedure below.

\subsubsection{Simulated Manipulator Pushing}
We used PyBullet \cite{coumans2019} to generate data from a simulated manipulator pushing task. 
We collected grayscale images of size $64 \times 64$ pixels, $\mathbf{x}_{t}^{1} \in \mathbb{R}^{64 \times 64}$, with pixel intensities rescaled to be in the range of zero to one.
The proprioceptive data consisted of the Cartesian position and velocity of the end-effector, while the haptic data included force and torque measurements along three axes.
We combined the haptic and proprioceptive data into a single second modality when both were used, $\mathbf{x}_{t}^{2} \in \mathbb{R}^{12}$, and otherwise used one of the two, $\mathbf{x}_{t}^{2} \in \mathbb{R}^{6}$. 
We argue that this is a reasonable decision since both proprioceptive and haptic data have similar characteristics, as opposed to, for example, image data. 
The control inputs were the end-effector velocity commands along the planar $x$ and $y$ directions, $\mathbf{u}_{t} \in \mathbb{R}^{2}$. 

A total of 4,800 trajectories were collected. 
The image data was recorded at a frequency of approximately 4 Hz and force-torque and proprioceptive data at a frequency of 120 Hz.
We concatenated sequences of measurements in order to keep a consistent number of time steps between modalities.
We used 4,320 trajectories for training and held out 480, or 10\%, for evaluation. 
Each trajectory was of length $T = 16$ (i.e., $16$ images and $16 \times 32$ (concatenated) force-torque and proprioception measurements in total).
The number of measurements per discrete time step was based on the respective frequency of each data source. 
The object being pushed was a single square plate. 
In order to collect training data, we used a policy with actions drawn from a fixed Gaussian distribution. 
We used the same initial object position for each trajectory.

\begin{figure}
	\vspace*{-4mm}
    \centering
    \captionsetup[subfigure]{font=footnotesize, justification=centering}
    \subfloat[Ellipse-1 on Delrin.]{\makebox[25mm][c]{
        \includegraphics{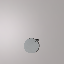}}
    }\hfil
    \subfloat[Ellipse-2 on plywood.]{\makebox[25mm][c]{
        \includegraphics{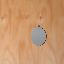}}  
    }\hfil
    \subfloat[Ellipse-3 on ABS.]{\makebox[25mm][c]{
        \includegraphics{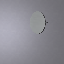}} 
    }\hfil
    \subfloat[Ellipse-3 on polyurethane.]{\makebox[25mm][c]{
        \includegraphics{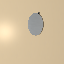}}
    }
    \caption{Downsampled images generated from the real-world MIT pushing dataset using existing tools \cite{Kloss_2020}. Realistic lighting and material textures are also rendered.}
    \label{fig:rendered_img}
	\vspace{-10mm}
\end{figure}

\subsubsection{MIT Pushing}
The MIT pushing dataset \cite{yu2016push} consists of high-fidelity real robot pushes carried out on various material surfaces and with various object shapes. 
Our dataset was a subset of trajectories with three different ellipse-shaped plates and four different surface materials (Delrin, plywood, ABS, and polyurethane). We followed the experimental protocol of the authors in \cite{lee2020}.

We used the code provided by previous work \cite{Kloss_2020} to preprocess the data and to artificially render the respective images of the trajectories (since no image data was collected as part of the original dataset). 
In Figure \ref{fig:rendered_img}, we show four examples of rendered images based on the real-world pushing data (i.e., object pose, end-effector pose, and force-torque data). 
We note that the rendered images are not completely realistic (e.g., the images are not occluded and only the manipulator's tip is rendered). 
However, for our purposes, the dataset provided an adequate starting point for preliminary experiments and comparisons.  

We downsampled the images to $64 \times 64$ pixels and transformed them to grayscale, $\mathbf{x}_{t}^{1} \in \mathbb{R}^{64 \times 64}$, with pixel intensities rescaled to be in the range of zero to one.
The proprioceptive data consisted of the $x$ and $y$ Cartesian coordinates of the end-effector.
The haptic data consisted of force measurements along the $x$ and $y$ (in-plane) axes and the torque measurements about the $z$ axis.
We combined the haptic and proprioceptive data to form a single second modality, $\mathbf{x}_{t}^{2} \in \mathbb{R}^{5}$, if both were used, and otherwise used one or the other, $\mathbf{x}_{t}^{2} \in \mathbb{R}^{3}$ or $\mathbf{x}_{t}^{2} \in \mathbb{R}^{2}$. 
The control inputs were position commands along the $x$ and $y$ axes, $\mathbf{u}_{t} \in \mathbb{R}^{2}$. 
The images were recorded at a frequency of $18$ Hz and the force and proprioceptive data were recorded at a frequency of $180$ Hz. 
We used a subset of 2,332 trajectories out of the total dataset (2,099 trajectories for training and 233, or approximately 10\%, held out for evaluation).
Each trajectory was of length $T=48$ (i.e., $48$ images and $48 \times 10$ (concatenated) force-torque and proprioception measurements in total).  
The data were collected using several pre-planned pushes with varying velocities, accelerations, and contact points and angles. 

\subsection{Network Architecture and Training}
\label{nat}
We parametrized our models with neural networks.
For the image encoder, we used a fully convolutional neural network based on the architecture of \cite{Ha2018-ho}. 
The respective decoder was a matching deconvolutional network.
For the proprioceptive and force-torque data, we used a simple 1D convolutional architecture for the encoder and a 1D deconvolutional network for the decoder. 

Our transition function or dynamics model, $p_{\mathbf{\theta}}(\mathbf{z}_{t} | \mathbf{z}_{t-1}, \mathbf{u}_{t-1})$ was parameterized as a single-layer GRU network \cite {cho2014properties} with 256 units that produced linear transition matrices (i.e., $\mathbf{z}_{t} = \mathbf{A}_{t} \mathbf{z}_{t-1} + \mathbf{B}_{t} \mathbf{u}_{t-1}$, where $\mathbf{A}_{t}$ and $\mathbf{B}_{t}$ are outputs of the network), as done in previous work \cite{Karl2016-ok, FraccaroKPW17, 2020_Limoyo_Heteroscedastic}.
During training, we sampled mini-batches of 32 trajectories. 
We applied weight normalization \cite{Salimans2016WeightNorm} to all of the network layers except for the GRU.
We used ReLU activation functions for all of the networks.
Our latent space for all experiments consisted of 16-dimensional Gaussians with diagonal covariance matrices.
We applied the \texttt{Adam} optimizer \cite{KingmaB14} with a learning rate of .0003 and a gradient clipping norm of 0.5 for the GRU network.

\subsection{Image Prediction Experiments}
\label{ipe}
\begin{figure}
    \centering
    \subfloat[Results from the synthetic dataset.\label{fig:quant_sim_graph}]{
        \includegraphics[width=0.75\columnwidth]{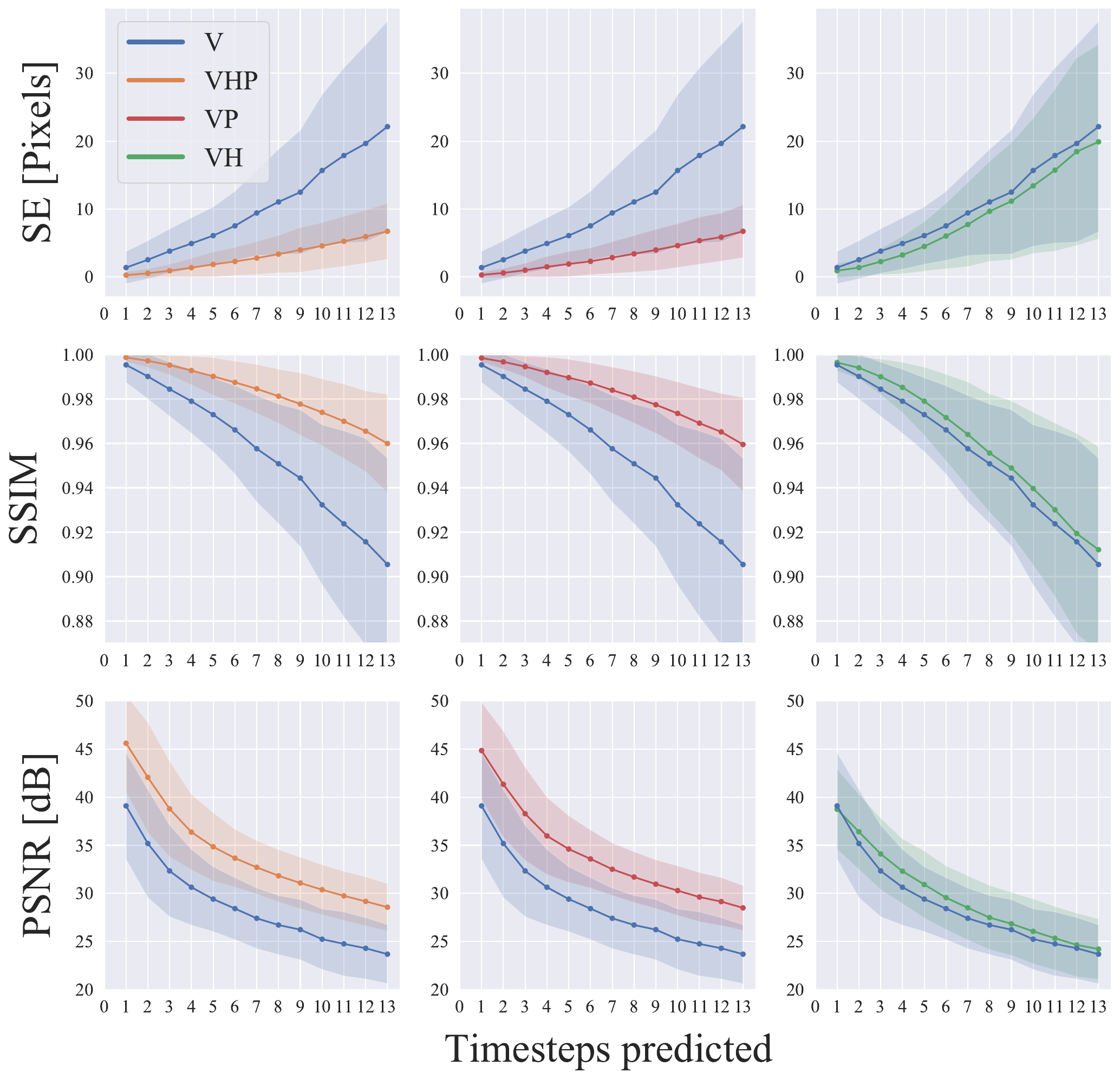}    
    }\hfil
    \subfloat[Results from the MIT dataset.\label{fig:quant_real_graph}]{
        \includegraphics[width=0.75\columnwidth]{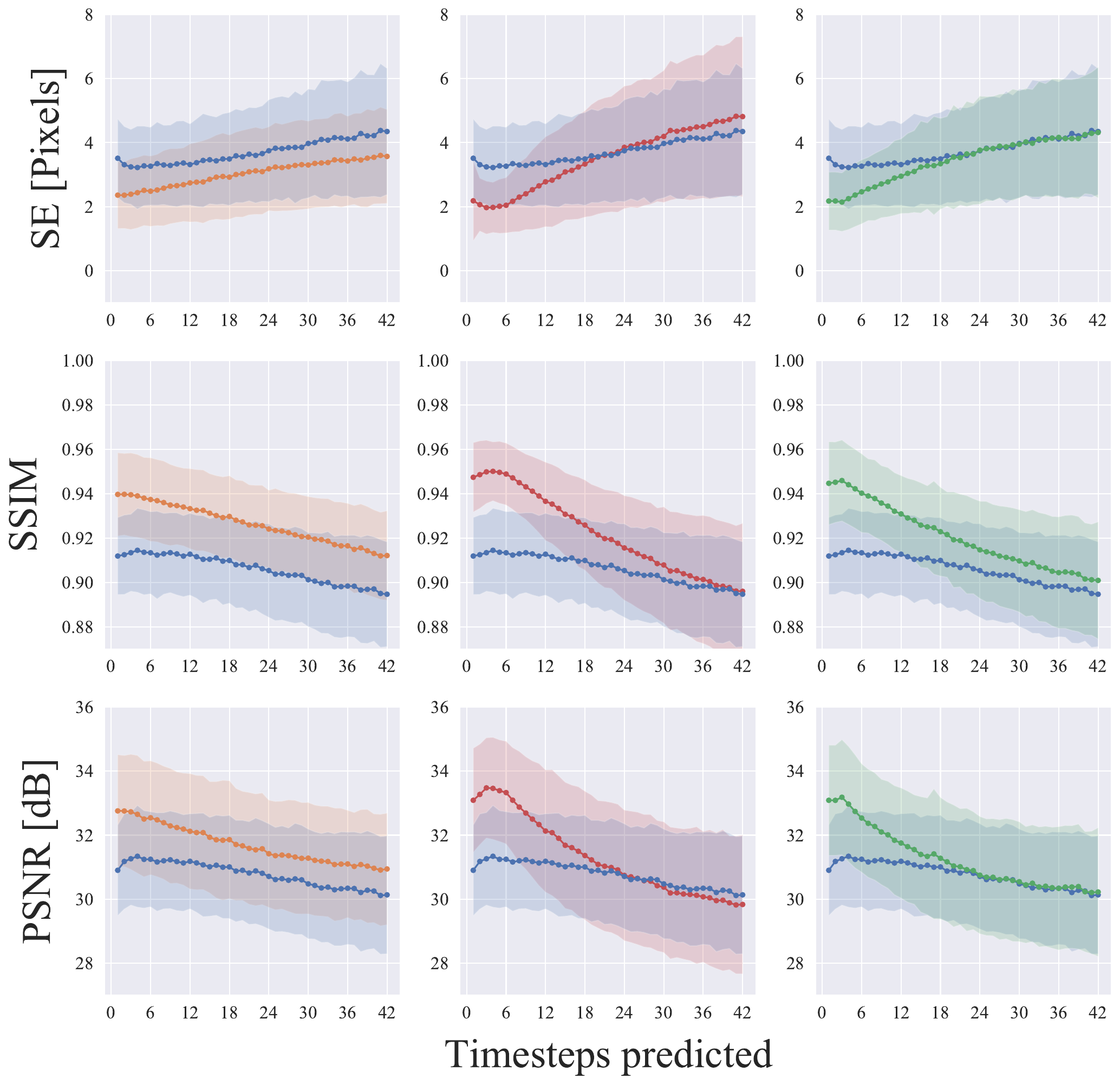}    
    }
    \caption{Graph of prediction quality over multiple prediction horizons from both datasets. We compare each model to the baseline vision-only (V) model. We plot the mean values of the SE, SSIM and PSNR with one standard deviation shaded.}
\end{figure}
In order to compare the models, we first evaluated the quality of the generated image sequences relative to the known ground truth images as a proxy for prediction quality. 
We began by encoding an initial amount of data into the latent space and then predicted future latent states with our learned dynamics models and known control inputs. 
Given these predicted states, we then decoded and generated future predicted images. 
We computed the squared error (SE) in terms of pixels, structural similarity index measure (SSIM), and peak signal-to-noise ratio (PSNR) between the ground truth images and predicted images. 
Accurate predictions translate into a lower SE, higher SSIM, and higher PSNR.

For the synthetic data, we first conditioned on four frames and predicted the next 11 frames in the sequence. 
Figure \ref{fig:quant_sim_graph} is a visualization of the mean score, with one standard deviation shaded, based on all of the held out test data. 
For clear comparisons, we overlay each multimodal model (i.e., VHP, VP, and VH) on top of the baseline vision-only model (i.e., V). 
\begin{table}
	\centering
	\caption{Quantitative prediction quality for the synthetic dataset. 
	We compared the predicted images' quality with their respective ground truth for four models trained with various subsets of modalities. 
	We show one standard deviation for the SSIM and PSNR values and calculate the average score across the entire predicted horizon.}
	\setlength{\tabcolsep}{16pt}
	\label{tab:quant_sim}
	\begin{tabular}{llll}
	\toprule
	\multicolumn{1}{c}{\textbf{Model}} & \multicolumn{1}{c}{\textbf{RMSE} ($\downarrow$)}       & \multicolumn{1}{c}{\textbf{SSIM} ($\uparrow$)}              & \multicolumn{1}{c}{\textbf{PSNR} ($\uparrow$)}    \\ \midrule
	V     & 3.243          & 0.955 $\pm$ 0.04          & 28.639 $\pm$ 5.84          \\ 
	VP    & 1.758          & 0.982 $\pm$ 0.02          & 33.935 $\pm$ 5.93          \\ 
	VH    & 2.956          & 0.961 $\pm$ 0.04          & 29.684 $\pm$ 5.70          \\ 
	VHP-C   & 2.911 & 0.962 $\pm$ 0.04 & 29.700 $\pm$ 5.39 \\ 
	VHP   & \textbf{1.749} & \textbf{0.983 $\pm$ 0.02} & \textbf{34.202 $\pm$ 6.19} \\ 
	\bottomrule
	\end{tabular}	
\end{table}
\begin{table}
	\centering
	\caption{Quantitative prediction quality with the MIT dataset as described in Table \ref{tab:quant_sim}.}
	\setlength{\tabcolsep}{16pt}
	\label{tab:quant_real}
	\begin{tabular}{llll}
	\toprule
	\multicolumn{1}{c}{\textbf{Model}} & \multicolumn{1}{c}{\textbf{RMSE} ($\downarrow$)}       & \multicolumn{1}{c}{\textbf{SSIM} ($\uparrow$)}              & \multicolumn{1}{c}{\textbf{PSNR} ($\uparrow$)}    \\ \midrule
	V     & 1.917         & 0.907 $\pm$ 0.02          & 30.806 $\pm$ 1.66          \\ 
	VP    & 1.850          & 0.923 $\pm$ 0.03          & 31.395 $\pm$ 2.28        \\ 
	VH    & 1.839          & 0.922 $\pm$ 0.03          & 31.298 $\pm$ 2.03          \\ 
	VHP-C    & 1.846          & 0.923 $\pm$ 0.03          & 31.333 $\pm$ 2.12          \\ 
	VHP   & \textbf{1.734} & \textbf{0.927 $\pm$ 0.02} & \textbf{31.741 $\pm$ 1.84} \\ 
	\bottomrule
	\end{tabular}
\end{table}
Additionally, we compile the average score over the entire predicted horizon and summarize the results in Table \ref{tab:quant_sim}.
We also include the baseline results of simply concatenating all modalities, labelled as VHP-C.

Overall, the VHP model performed the best for all three metrics when averaged over all prediction lengths, but it was closely followed by the VP model. 
All multimodal models outperformed the baseline, vision-only model, although the improvement in the case of the VH model was marginal.
We hypothesize that this result may be due to the quality of the simulated force torque sensor data and the simulated contact dynamics.
In this case, proprioception was the most effective second modality, as demonstrated by the performance of the VP model. 
However, using vision, haptic and proprioceptive data together with the VHP model led to the best results.

Notably, simply concatenating each modality (i.e., VHP-C) yielded poorer performance when compared to our product of experts approach (i.e., VHP).
The product of experts appears to have an appealing `filtering' inductive bias for prediction problems. 
The product of experts formulation enables decisions to be made about when to use each modality at each time step based on the respective uncertainties. 
Further, each expert can selectively focus on a few dimensions without having to cover the full dimensionality of the state. 
Finally, a product of experts produces a sharper final distribution than the individual expert models \cite{poe2002}. 
This idea is well known in the context of digit image generation: one low-resolution model can generate the approximate overall shape of the digit while other local models can refine segments of the stroke with the correct fine structure \cite{poe2002}.
\begin{figure}
	\centering
	\includegraphics[width=0.915\textwidth]{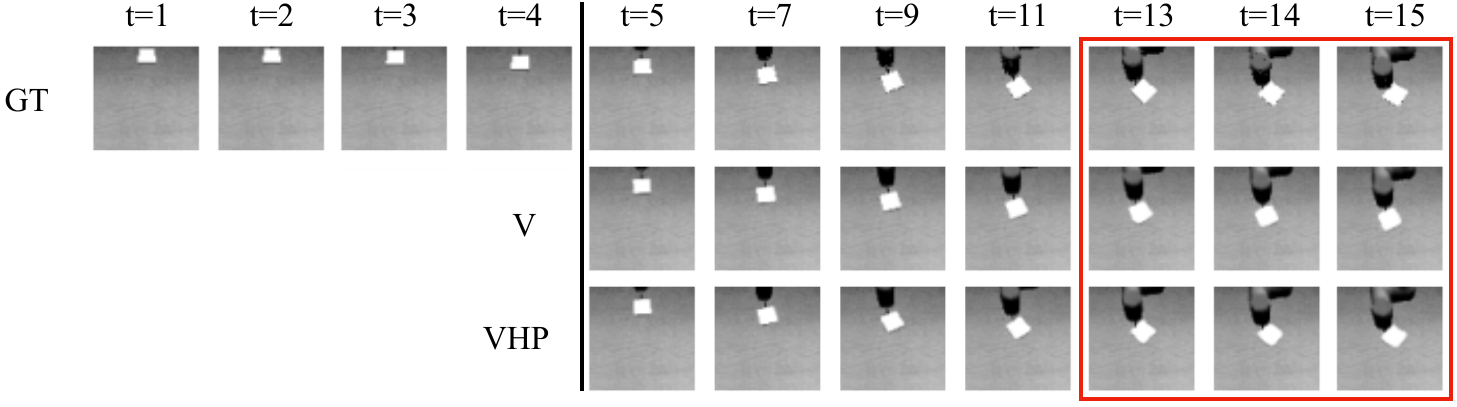}
	\caption{A prediction example that demonstrates a failure mode of the vision-only (V) model trained on the synthetic dataset; the model fails to fully predict the object's true motion (GT) and instead predicts less counterclockwise rotation in the later frames (denoted by red box). In contrast, the multimodal model (VHP) correctly predicts the object's motion.}
	\label{fig:qual_sim}
	\vspace{-2mm}
\end{figure}
\begin{figure}
	\centering
	\includegraphics[width=0.915\textwidth]{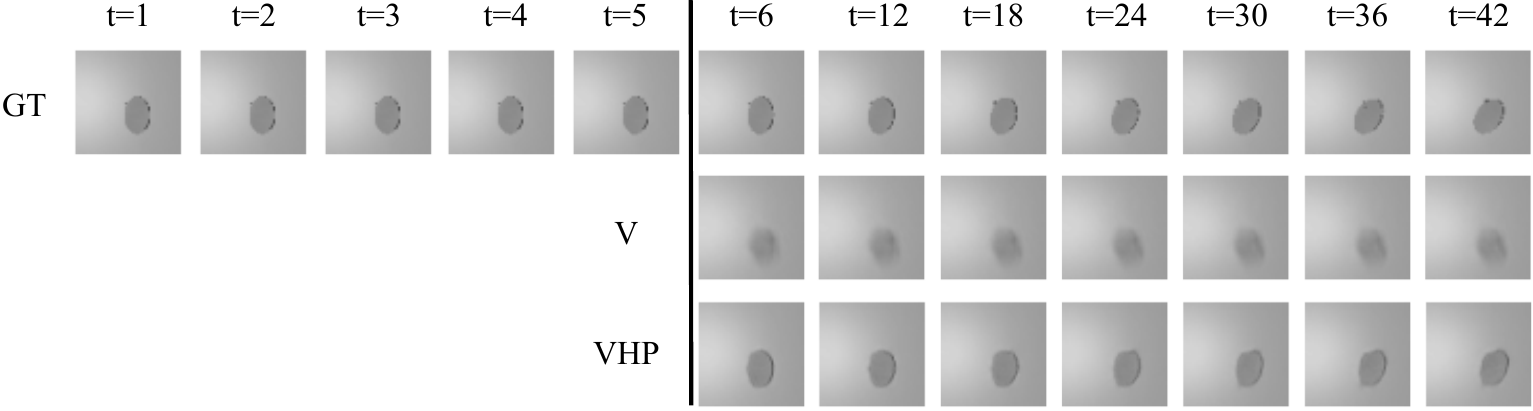}
	\caption{A prediction example that demonstrates a different failure mode of the vision-only (V) model trained on the MIT dataset; the model is unable to produce `crisp' predictions of the object form and instead outputs a blurry average at all time steps. The multimodal model (VHP) is able to predict the slight angle of the ellipse (clockwise), matching the ground truth (GT).}
	\label{fig:qual_real}
	\vspace{-5mm}
\end{figure}
Figure \ref{fig:qual_sim} is a visualization of a selected roll-out or prediction that demonstrates a typical failure mode of the vision-only model. 
As shown by the later frames generated by the vision-only model in the red box outline, the overall motion of the object is almost correct, but the finer-scale changes (e.g., the exact amount of rotation) are not well captured.
Our VHP model, however, was able to capture these smaller changes.

We ran similar image prediction experiments for the MIT pushing dataset.
In this case, we first conditioned on five frames and predicted the next 37 frames in the sequence. 
As shown by Figure \ref{fig:quant_real_graph}, both the VP and VH models performed slightly better than the full VHP during the early time steps according to all three metrics (SE, SSIM and PSNR). 
However, over a longer prediction horizon the VHP model was clearly superior.
For the real-world data, unlike in the synthetic case, the VH model significantly outperformed the baseline V model. 
Table \ref{tab:quant_real} lists the average scores over the entire prediction horizon. 
Consistent with the previous synthetic experiment, simply concatenating the modalities (VHP-C) led to poorer performance when compared to the product of experts model (VHP).
Each multimodal model performed better than the baseline vision-only model.
We visualize another failure mode of the vision-only model in Figure \ref{fig:qual_real}. 
In this case, we observe that the vision-only model failed to capture the more subtle dynamics of the object, which ended up angled and tilted clockwise; the model defaulted to outputting a blurry average at the approximate location of the object. 
On the other hand, our VHP model was able to produce a relatively crisp and accurate prediction of the object pose.

\subsection{Regression Experiments}
We further evaluated the predictive capability of the models with a downstream regression experiment that measured how well the model was able to capture and predict the underlying position of the object being pushed.
As is commonly done in the self-supervised representation learning literature \cite{grill2020bootstrap, kornblith2019better}, we trained separate regressors on top of the frozen representations and dynamics models to do this.
\begin{figure}
    \centering
    \subfloat[Results from the synthetic dataset.\label{fig:ols_sim}]{
        \includegraphics[height=2.00in]{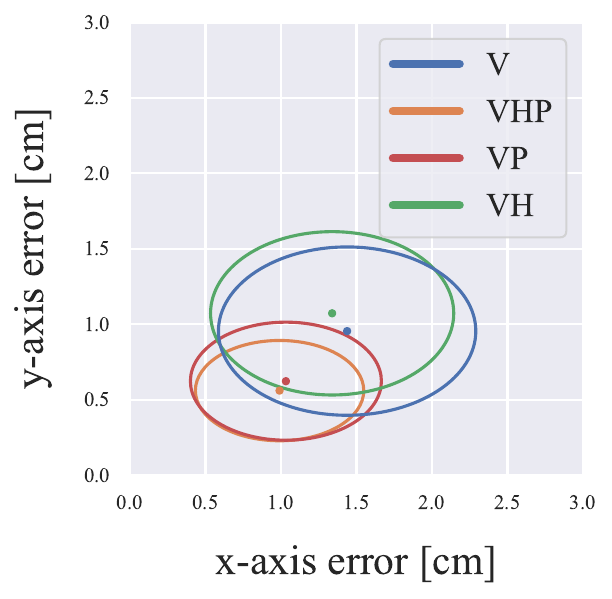}    
    }
    \subfloat[Results from the MIT dataset.\label{fig:nn_real}]{
        \includegraphics[height=2.00in]{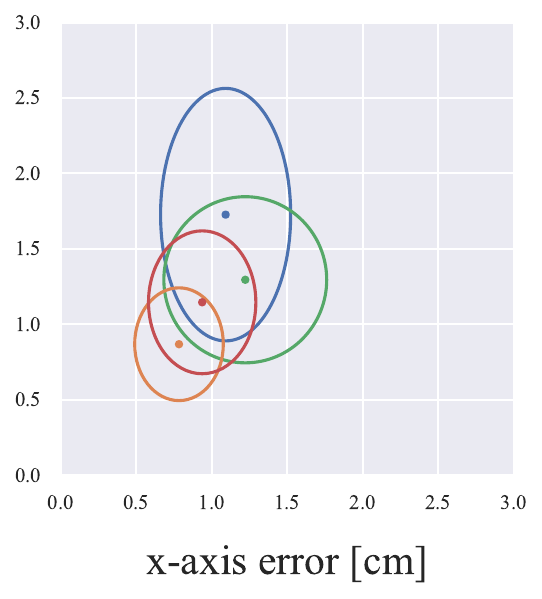}    
    }
    \caption{Mean translation errors for regressing the object's $x$ and $y$ coordinates from the predicted latent states. The ellipses represent one standard deviation.}
	\label{fig:error_ellipses}
	\vspace{-5mm}
\end{figure}

We first encoded an initial latent state $\mathbf{z}_{i}$ (i.e., a filtered state) from a set of observations $\mathbf{x}_{1:i}$. Then, using our learned dynamics and $h$ known future controls $\mathbf{u}_{i:i + h}$, we predicted $h$ future latent states $\hat{\mathbf{z}}_{i + 1:i + 1 + h}$ (i.e., predicted states).
Using the predicted states as inputs, we regressed the ground truth positions of the object, while keeping the weights of all of our previously-learned networks frozen.
Poor regression results would indicate lower correlation between the predicted latent states and the object's ground truth position.
This in turn would imply that the learned latent representation did not encode all of the necessary information to represent the state of the object, or that the learned dynamics of the object were inaccurate, or both.
We kept the same prediction horizons from the image prediction experiments in Section \ref{ipe}.

For our results with synthetic data, we found ordinary least squares to be adequate for regression.
Figure \ref{fig:ols_sim} shows that the mean and variance of the absolute translation errors from the regression are lowest for the full VHP model. 
We (again) observe that the inclusion of haptic data alone (VH) as an extra modality in the model did not lead to huge improvements.
On the other hand, using proprioception (VP) did lead to significant improvements.
The inclusion of both proprioceptive and haptic data (VHP) outperformed using proprioceptive data alone (VP).

For the MIT pushing dataset, we trained a separate simple neural network with a single hidden layer of size 50 to regress the position of the object. 
The real-world results generally match our synthetic results, as shown by Figure \ref{fig:nn_real}. 
Including any sort of multimodal data reduced the mean and variance of the absolute translation errors when compared to the results from the vision-only model. 

As an additional benchmark with the MIT dataset, we compared the accuracy of our regressed values of the position of the object to those computed by a related approach on differentiable filtering \cite{lee2020}. 
Using the same dataset, the authors trained a variety of differentiable filters to directly regress the positions of the objects from the multimodal sensor data in a supervised manner using ground truth labels.
Because the multimodal differentiable filters were trained with ground truth annotations of the objects' locations, while our model is completely self-supervised, the results provide a reasonable estimate of an upper bound on the expected performance.
Table \ref{tab:rmse_comparison} demonstrates that our regression results, from both the predicted and filtered states, are comparable.
\begin{table}[]
  \centering
  \caption[]{Root-mean-square error (RMSE) of the objects' regressed positions with the MIT pushing dataset. We compared both the filtering (filt.) and prediction (pred.) results from our self-supervised multimodal model (VHP) to various types of differentiable filters from \cite{lee2020}.}
  \setlength{\tabcolsep}{16pt}
  \label{tab:rmse_comparison}
  \begin{tabular}{@{}lc@{}}
  \toprule
  \multicolumn{1}{c}{\textbf{Model}}                           & \textbf{RMSE} {[}\textbf{cm}{]} ($\downarrow$) \\ \midrule
  EKF (Sup. w/ GT, filt.) & 1.33          \\
  PF (Sup. w/ GT, filt.)  & 1.14          \\
  LSTM  (Sup. w/ GT, filt.)          & 2.32          \\
  VHP (Self-sup., filt.) (Ours)         & 1.80          \\
  VHP (Self-sup., pred.) (Ours)                  & 1.96           \\ \bottomrule
  \end{tabular}
  \vspace{-10mm}
\end{table}

\section{Conclusions \& Future Work}
\label{con}
We have presented a self-supervised sequential latent variable model for multimodal time series data.
Our probabilistic formulation extends existing latent dynamics models, a key backbone of many methods in vision-based control and model-based reinforcement learning, to multimodal sensor data.
We provided a case study of a manipulator pushing task where visual, haptic, and proprioceptive data streams were available.
For latent dynamics in particular, we demonstrated that a principled probabilistic formulation for fusing sequential multimodal data performed significantly better than the common baseline of directly concatenating each modality in the latent space.
Additionally, our learned self-supervised approach was shown to be competitive with an existing multimodal differentiable filtering method that relies on supervised ground truth labels.
As interesting avenues for future work, we plan to further investigate the properties of the learned uncertainties used for multimodal fusion and to use our model for control policy learning.

\bibliographystyle{splncs}
\bibliography{robotics_abbrv, references}

\end{document}